\documentclass{article}
\usepackage{ijcai22}

\usepackage{times}
\usepackage{soul}
\usepackage{url}
\usepackage[hidelinks]{hyperref}
\usepackage[utf8]{inputenc}
\usepackage[small]{caption}
\usepackage{subcaption}
\usepackage{graphicx}
\usepackage{amsmath}
\usepackage{amsthm}
\usepackage{booktabs}
\usepackage{algorithm}
\usepackage{algorithmic}
\usepackage{xcolor}
\usepackage{dsfont}

\usepackage{amsmath}
\usepackage{amssymb}

\newtheorem{theorem}{Theorem}
\newtheorem{lemma}{Lemma}
\newtheorem{definition}{Definition}
\newtheorem{corollary}{Corollary}[theorem]

\title{Adaptive Information Belief Space Planning}

\author{
Moran Barenboim$^1$
\and
Vadim Indelman$^2$
\affiliations
$^1$Technion Autonomous Systems Program\\
$^2$Department of Aerospace Engineering\\
$^{^{^{\ }}}$Technion - Israel Institute of Technology, Haifa 32000, Israel
\emails
\small moranbar@campus.technion.ac.il,
\small vadim.indelman@technion.ac.il
}

\begin{document}
\maketitle

\begin{abstract}
    Reasoning about uncertainty is vital in many real-life autonomous systems. However, current state-of-the-art planning algorithms cannot either reason about uncertainty explicitly, or do so with a high computational burden. Here, we focus on making informed decisions efficiently, using reward functions that explicitly deal with uncertainty. We formulate an approximation, namely an abstract observation model, that uses an aggregation scheme to alleviate computational costs. We derive bounds on the expected information-theoretic reward function and, as a consequence, on the value function. We then propose a method to refine aggregation to achieve identical action selection with a fraction of the computational time.
\end{abstract}

\section{Introduction}
Planning under uncertainty is a recurrent aspect in many practical autonomous systems. The uncertainty stems from the fact that in many real-life scenarios the state of the agent and the surrounding world is not known precisely. Some of the most common sources for uncertainty include sensor noise, modeling approximations and the unknown environment of the agent. A common way to decision making while considering uncertainty is to formulate the problem as a partially observable Markov decision process (POMDP). 

Since in POMDPs states are unobserved directly, all past action-observation pairs, coined history, serve as an alternative for the decision making. Accumulating all the history over long trajectories may be expensive to memorize. Instead, a common approach is to calculate a distribution over the unobserved states, also known as belief states. In most POMDP formulations, belief states are assumed to be sufficient statistics, such that they are equivalent to the history when considering decision making \cite{Thrun05book}. An optimal solution to the POMDP is a policy that maximizes some objective function, usually defined as the sum of expected future reward values over the unknown states. A policy then maps each belief state to an action. Unfortunately, the exact solution of a POMDP is computationally infeasible for many practical POMDPs \cite{Papadimitriou87math}.

A reward in a POMDP is a function that commonly receives a state as an input, and maps each state to a scalar value. However, when  the reward receives a belief as an input, the problem formulation is considered an extension to POMDP, called $\rho$-POMDP \cite{Araya10nips}, POMDP-IR \cite{Spaan2015aamas} or Belief Space Planning (BSP) \cite{Platt10rss}, \cite{VanDenBerg12ijrr}, \cite{Indelman15ijrr}. An objective function on the distribution itself allows reasoning about the uncertainty directly, which arises in many problems such as active localization \cite{Burgard97ijcai}, information gathering \cite{Hollinger14ijrr} and active SLAM \cite{Kim14ijrr}. Although POMDP formulation allows to reason about uncertainty implicitly, it is insufficient for problems where the goal is defined over a distribution. For instance, consider the active localization problem, the goal of the agent is not to reach a certain destination, but to reduce uncertainty about its state. 

\begin{figure}
    \centering
    \begin{subfigure}{0.48\textwidth}
         \includegraphics[width=1\textwidth]{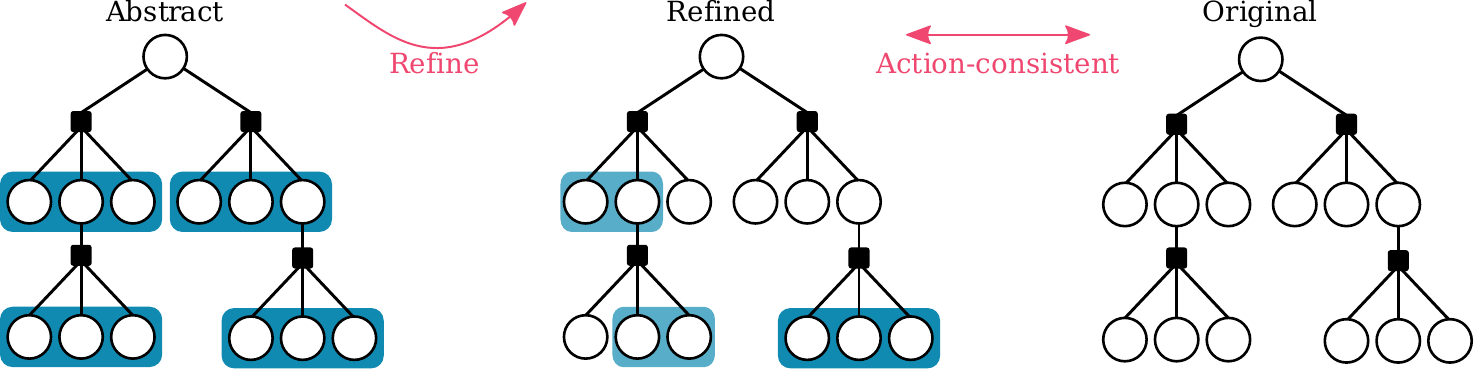}
    \end{subfigure}
    \caption{An illustration of our approach. The blue clusters correspond to a single evaluation of the reward function across different posterior nodes, which is faster to compute. Then, the algorithm initiates a refinement procedure; The refined clusters guarantee the same action selection as the original reward evaluation. }
    \label{fig:tree}
\end{figure}

Common approaches for measuring uncertainty are information-theoretic functions, such as entropy, information gain and mutual information. For continuous distributions, which are of interest in this paper, calculating the exact values of information-theoretic rewards involves intractable integrals in the general case. Thus, they are amenable to different approximations, such as kernel density estimation (KDE), Voronoi diagrams \cite{Miller03icassp} or sampling-based approximations \cite{Boers10fusion}. Unfortunately, all such approximations are expensive to compute, as they require quadratic costs in the number of samples and are usually the bottleneck of planning algorithms. 

Since finding an exact solution to a POMDP is intractable, approximate algorithms have been developed. Tree search algorithms, which are the main focus of this paper, are a prominent approach for such approximation; Instead of considering all possible belief states, tree search algorithms reduce the belief space to a reachable subset, starting from the prior belief node. In this paper, we consider an online paradigm, where instead of calculating a policy in advance, the planner needs to find an (approximately-) optimal action, by building a tree every time step. POMCP \cite{Silver10nips}, is a tree search algorithm, which extends MCTS \cite{Browne12ieee} to the POMDP framework; It is considered as a state-of-the-art (SOTA) algorithm, however, it is limited to a discrete state, action and observation spaces. DESPOT \cite{Somani13nips} and its descendants \cite{Ye17jair,Garg19rss} consider $\alpha$-vectors in their derivations and are thus limited to reward functions that are linear in the belief. However, information-theoretic functions are usually not linear with respect to the belief and thus not supported by these approaches.

POMCPOW \cite{Sunberg18icaps} is an extension of the POMCP algorithm, to deal with continuous action and observation spaces. In POMCPOW, each belief node is represented by a set of particles. The number of particles in each node is determined by the number of traversals within this node in planning. Due to the exploratory nature of the algorithm, most belief nodes contain only a few particles, which are unsuitable to approximate belief-dependent reward. In PFT-DPW algorithm \cite{Sunberg18icaps}, each expanded node in the belief tree contains the same number of particles and is better suited for belief-dependent rewards. A recent result by \cite{Hoerger21icra} shows comparable or better performance than current SOTA algorithms, by considering trajectories of particles and using a weighted particle filter to extract trajectories for any observation, but suffers from the same problem as POMCPOW. \cite{Flaspohler19ral} uses MCTS with information-theoretic as heuristic over state-dependent objective function for information gathering. 

More closely related to our work is \cite{Thomas21arxiv}, which interleave MCTS with $\rho$-POMDP. Their approach considers a discrete observation space. In each traversal of the tree, their algorithm adds a fixed set of particles propagated from the root node, which results in an increased number of samples at each node as the algorithm progresses. According to the authors, the main motivation is a reduced asymptotic bias. Given a time budget, a significant reduction in the number of nodes explored is observed by the authors of \cite{Thomas21arxiv}, which in turn impacts the quality of the policy. \cite{Fischer20icml} consider a belief-dependent reward, in which they build upon PFT-DPW \cite{Sunberg18icaps}. Instead of maintaining the same particle set in each posterior node, they reinvigorate particles in every traversal of the posterior node. Then, they propose to average over different estimations of the reward function. \cite{Fischer20icml} suggest estimating the reward function using KDE, which scales quadratically with the number of state samples. \cite{Sztyglic21arxiv} consider a sampling-based approximation to evaluate differential entropy. They propose a simplification procedure that alters the number of state samples and prune sub-optimal action branches, using bounds relative to the non-simplified estimator. In a follow-up work, \cite{Sztyglic21arxiv_c} propose an approach to interleave simplification with MCTS while maintaining tree-consistency, thereby increasing computational efficiency. Our work is complementary to these works and can be combined. 

In this paper, we show an approach to alleviate the computational burden of calculating reward at each belief node of the tree, while guaranteeing an identical solution. We focus on Shannon's entropy and differential entropy, alongside state-dependent reward functions. Our approach relies on clustering different nodes and evaluating an approximated belief-dependent reward once on this entire cluster, that is, all the nodes within a cluster share the same reward value. As a result, the estimated value function is affected. To relate the approximated value function to the one that would originally be calculated, each node maintains a lower and upper bound on the value function.

Consequently, our main contributions are as follows. First, we introduce an abstract observation model. We use the model to form an abstraction of the expected reward, namely, a weighted average of state-dependent reward and entropy. Then, using the abstract model, we show how computational effort is alleviated. Second, we derive deterministic lower and upper bounds for the underlying expected reward values and the value function. Third, we introduce a new algorithm, which is able to tighten the bounds upon demand, such that the selected action is guaranteed to be identical to the non-simplified algorithm. Last, we evaluate our algorithm in an online planning setting and show that our algorithm outperforms the current state-of-the-art.

\section{Preliminaries}
\subsection{POMDP Formulation}
POMDP is defined as a 7-tuple, $(S,A,O,T,Z,R,b_0)$, where, $S,A,O$ denotes the state, action and observation spaces respectively. $T(s'\mid s,a)$ is the transition density function, $Z(o \mid s)$ is the observation density function and $b_0$ denotes the initial belief distribution. A history, $H_t$, is a shorthand for all past action-observation sequences $(a_0, o_1, \dots, a_{t-1}, o_t)$ up to current time. We use $H_t^-$ to denote the same history, without the last observation, i.e. $H_t^- = (a_0, o_1, \dots, a_{t-1})$. Similarly, $b_t^-$ is a belief conditioned on $H_t^-$. 

In this work, we focus on a reward function defined as a weighted sum of state-dependent reward and entropy,
\begin{equation} \label{def:reward}
R(b,a,b') = \omega_1\mathbb{E}_{s\sim b'}\left[r( s ,a)\right] + \omega_2\mathcal{H}(b'),
\end{equation}
where $b'$ is the subsequent belief to $b$ and $\mathcal{H}(\cdot)$ is either differential entropy or Shannon's entropy. The dependence of \eqref{def:reward} on both $b$ and $b'$ stems from the definition of the differential estimator, as will be shown in Section \ref{subsec:ContObs}. A policy, $\pi(\cdot)$, maps a belief to an action to be executed. Given a belief at time $t$, each policy corresponds to a value function, 
\begin{equation}
V^{\pi }( b_{t}) = \mathbb{E}_{o}\left[\sum\nolimits _{\tau=t} ^{\mathcal{T}-1} R( b_{\tau} ,\pi_{\tau}(b_{\tau}), b_{\tau+1})\right],
\end{equation}
which is the expected cumulative reward following the policy, $\pi$. Similarly, an action-value function,
\begin{equation}
Q^{\pi }( b_{t} ,a_{t}) = \mathbb{E}_{o_{t+1}}\left[R( b_{t} ,a_{t}, b_{t+1}) +  V^{\pi }( b_{t+1})\right],
\end{equation}
is the value of executing action $a_t$ in $b_t$ and then following the policy $\pi$.

\subsection{Belief-MDP}
A belief-MDP is an augmentation of POMDP to an equivalent MDP, by treating the belief-states in a POMDP as states in the Belief-MDP. Subsequently, algorithms developed originally for MDPs can be used for solving POMDPs or BSP problems with slight modifications, a property that we exploit in this paper.

\section{Expected Reward Abstraction}
In this section we introduce the notion of an abstract observation model and show how this model can be utilized to ease the computational effort. We then derive bounds on the expected reward and, as a consequence, on the value function.

We start this section with a theoretical derivation, where we assume the reward can be calculated analytically. This appears in special cases, e.g. when the reward is solely entropy and the belief is parametrized as a Gaussian or when the state space is discrete. In this part we assume the observation space is discrete and thus the expected reward can be calculated analytically. In the second part of this section, we relax those assumptions. Generally, the observation and state spaces can be continuous and the belief may be arbitrarily distributed. To deal with such cases, we derive an estimator of the expected reward, in which the belief is approximated by utilizing a particle filter.  

\begin{figure}[t]
    \centering
    \begin{subfigure}{0.48\textwidth}
         \includegraphics[width=1\textwidth]{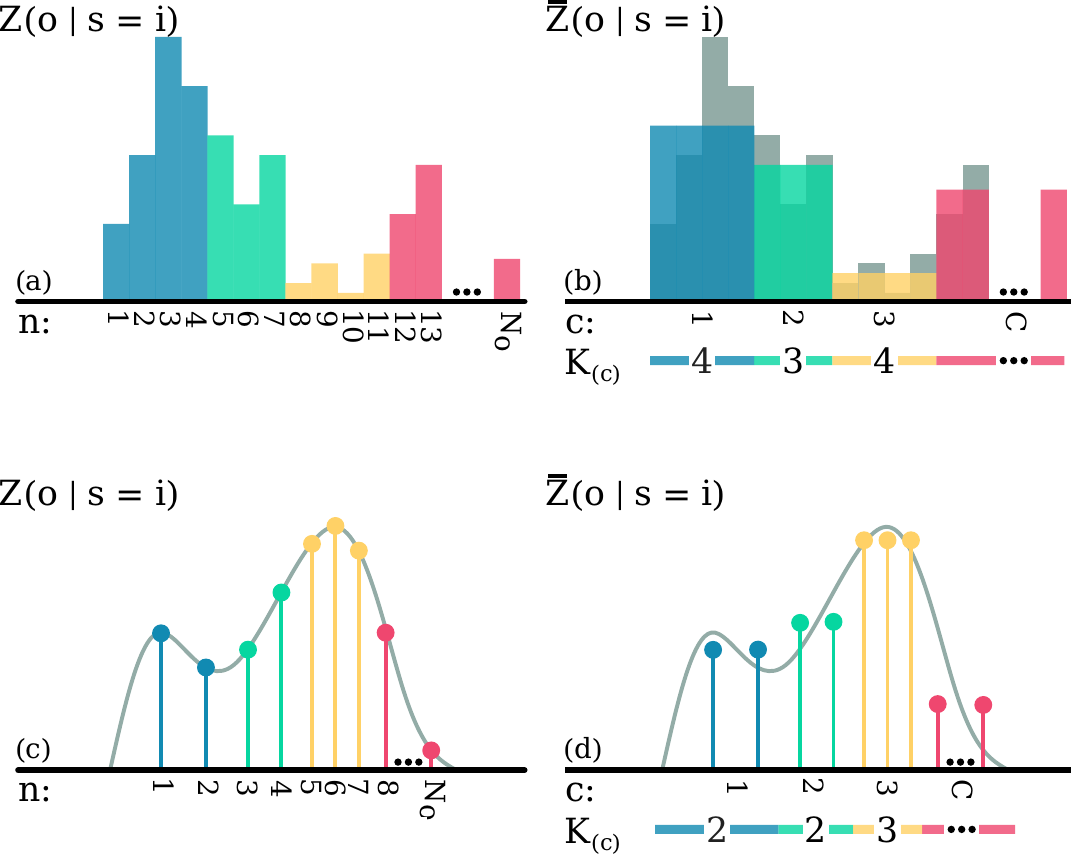}
    \end{subfigure}
    \caption{An abstraction of the observation model: (a) original discrete observation model with $N_o$ observations. (b) Abstract discrete observation model with $C$ clusters. (c) Sample set from the original continuous observation model, with $N_o$ observations. (d) Abstract sample set with $C$ clusters.}
    \label{fig:abs}
\end{figure}

For both the discrete and continuous observation spaces, we present an abstract observation model,
\begin{definition}[Abstract observation model] \label{definition1}
An abstract observation model assigns a uniform probability to all observations within a single set,
\begin{equation}\label{eq:abst}
    \bar{Z}( o^{j}\mid s) \doteq \frac{1}{K}\sum _{k=1}^{K} Z\left( o^{k} \mid s\right) \ \ \ \ \ \forall j \in [1,K]
\end{equation}
\noindent where $Z\left( o^{k} \mid s\right)$ corresponds to the original observation model over different observation realizations, $o^{k}$, and $K$ denotes the cardinality of observations within that set.
\end{definition}
\noindent The abstract model aggregates $K$ different observations and replaces the original observation model when evaluating the reward, see Figure \ref{fig:abs}. In the continuous case we revert to observation samples, thus the summation in \eqref{eq:abst} corresponds to different observation samples. A more precise explanation on the continuous case will be given in Section \ref{subsec:ContObs}.

\subsection{Discrete Observation Space}
Since calculating the exact value of a reward function in every belief node is expensive, we now formulate an approach to evaluate reward once for an entire set of $K$ posterior beliefs. Such an abstraction results in a decreased number of reward evaluations in planning. We show that when constructing the aggregation scheme as a uniform distribution over a set of observations, one can achieve tight upper and lower bounds on the expected entropy, defined as $-\mathbb{E}\left[\sum_{s}b(s)log(b(s))\right]$. Moreover, we show that abstraction for the expected state-dependent reward does not affect its value, which remains identical with and without abstraction. 

Denote the cardinality of observation space with $N_o$, we partition the observations to $C$ clusters and denote the number of observations within each cluster as $K$, see Figure \ref{fig:abs}. Generally, each cluster may contain a different number of observations, $N_{o} =\sum _{c=1}^{C} K( c)$; For clarity, we assume $K$ is identical for all clusters, but the results below are easily extended to the more general case. 

Our \emph{key result}, stated in the lemma below, corroborates the intuition that utilizing an abstract observation model results  in a reduced number of reward evaluations.


\begin{lemma} \label{lemma1}
Evaluation of the expected reward with an abstract observation model, \eqref{eq:abst}, requires only $C$ evaluations of the reward, instead of $N_o$, where $C = \frac{N_o}{K}$. That is,
\begin{gather} \label{eq:lemma1}
\sum\nolimits _{n=1}^{N_{o}}\bar{\mathbb{P}}\left( o_{t+1}^{n} \mid H_{t+1}^{-}\right)
R\left( b_{t},a_{t},\bar{b}_{t+1}\right)= \\ 
K\sum\nolimits _{c=1}^{C}\bar{\mathbb{P}}\left( o_{t+1}^{c\cdot K} \mid H_{t+1}^{-}\right)
R\left( b_{t},a_{t},\bar{b}_{t+1}\right). \notag
\end{gather}
\end{lemma}
\begin{proof}
see appendix A.1.
\end{proof}
\noindent Here, $o_{t+1}^{c\cdot K}$ denotes a single representative observation of the cluster $c$ and
\begin{align}
&\bar{\mathbb{P}}\left( o^{n} \mid H^{-}\right) \doteq \sum_{s\in S} \bar{Z}(o^{n} \mid s) b^- , \label{eq:obs_exp} \\
&\bar{b} \doteq \frac{\bar{Z}(o^{n} \mid s) b^{-}}{\sum_{s^{'} \in S} \bar{Z}(o^{n} \mid s^{'}) b^{-}} \ .\label{eq:abs_belief}
\end{align} 
For the continuous state case, simply replace summations with integrals. Furthermore, the expected state-dependent reward remains unchanged when evaluated over abstract belief and expectation,
\begin{lemma} \label{lemma_state_theoretical}
The value of the expected state-dependent reward is not affected by the abstraction shown in \eqref{eq:abst}, i.e.,
\begin{gather}
\mathbb{E}_o\left[ \mathbb{E}_{s\sim b}\left[r\left( s,a\right)\right]\right]=
\bar{\mathbb{E}}_o\left[ \mathbb{E}_{s\sim \bar{b}}\left[r\left( s,a\right)\right]\right] . \label{eq:lemma_state_theoretical}
\end{gather}
\end{lemma}
\begin{proof}
see appendix A.2.
\end{proof}
\noindent Note that $\{\mathbb{E}_o,\mathbb{E}_{b}\}$ and $\{\bar{\mathbb{E}}_o,\mathbb{E}_{\bar{b}}\}$ correspond to expectations with the original and abstract observation models, \eqref{eq:obs_exp} and \eqref{eq:abs_belief}. 

We now transition to the main theorem of this paper. Using \eqref{eq:abst} as the abstraction mechanism, we show that,

\begin{theorem} \label{thm1}
The expected entropy is bounded from above and below by,
\begin{equation} \label{eq:mean_entropy_bounds}
0\leq \bar{\mathbb{E}}_{o}\left[\mathcal{H}\left( \bar{b}\right)\right] -\mathbb{E}_{o}\left[\mathcal{H}\left( b\right)\right] \leq log( K) .
\end{equation}
\end{theorem}
\begin{proof}
see appendix A.3.
\end{proof}
\noindent The difference between the expected entropy and the abstracted one is bounded from below by zero. Since the entropy evaluates the uncertainty, the interpretation of this result is quite intuitive; the uncertainty cannot reduce when using abstracted models. The upper bound depends on the number of observations we choose to abstract, $K$. When $K=1$, that is, each cluster contains a single observation, both the upper and lower bounds are zero and the abstract expected entropy equals the original expected entropy. Note that from \eqref{def:reward}, \eqref{eq:lemma_state_theoretical} and \eqref{eq:mean_entropy_bounds} it follows that,
\begin{equation} \label{eq:diff_exp_reward}
    0\leq \!{\bar{\mathbb{E}}}_{o}\left[R\left( {b},a,{\bar{b}}'\right)\right] \!-\!{\mathbb{E}}_{o}\left[R\left( {b},a,{b}'\right)\right] \leq \omega_2 log( K) .
\end{equation}
We now generalize those results and show that the value function is bounded. An abstract value function is defined as,
\begin{equation}
    \bar{V}^{\pi }( b_{t}) =  \bar{\mathbb{E}}_{o_{t+1}}\left[R(b_{t},\pi_{t}(b_{t}),\bar{b}_{t+1})\right] + \mathbb{E}_{o_{t+1}}\left[\bar{V}^{\pi }( b_{t+1})\right] . \label{eq:ent_val_func}
\end{equation}
As a direct consequence of \eqref{eq:diff_exp_reward} and \eqref{eq:ent_val_func},
\begin{corollary} \label{corrAnalyticallValueFunc}
The difference between the original value function and the abstract value function is bounded by, 
\begin{equation} \label{eq:thm_exact}
0 \leq \bar{V}^{\pi }( b_{t}) -V^{\pi }( b_{t}) \leq \mathcal{T}\cdot \omega_2 log( K).
\end{equation}
\end{corollary}
\begin{proof}
see appendix A.4.
\end{proof}
\noindent This result allows us to bound the loss when applying observation abstraction. In Section \ref{alg_sec}, we devise an algorithm that adapts the bounds so that the same best action will be chosen, with and without abstraction, while expediting planning time.

\subsection{Continuous Observation Space}\label{subsec:ContObs}
Planning with entropy as reward over a continuous observation space is more cumbersome as it requires calculation of,
\begin{gather} \label{exp_diff_ent}
\mathbb{E}[ \mathcal{H}( b_{t})] = 
-\int _{o_{t}} \mathbb{P}( o_{t} \mid H_{t}^{-})
\int _{s_{t}} b( s_{t}) log( b( s_{t})) .
\end{gather}
First, the probability density function of the belief may be arbitrary, so the integral over the state has no closed-form expression. Moreover, usually, there is no access to the density functions due to the difficulty of exact Bayesian inference, but only to samples. Second, even if the differential entropy could be evaluated, integrating over the observation space makes this calculation intractable. Consequently, it is only feasible to estimate the value function, here denoted by $\hat{V}^{\pi }( \cdot)$. To approximate the posterior belief at each time step, we employ the commonly used particle filter, see e.g. \cite{Thrun05book}. Inspired by derivations in \cite{Boers10fusion}, we estimate \eqref{exp_diff_ent} using state samples obtained via a particle filter. We then sample observations using the given observation model, conditioned on the state particles. This is a common procedure in tree search planning, see for example \cite{Sunberg18icaps}. Using the state and observation samples, we derive an estimator to \eqref{exp_diff_ent},
\begin{gather} \label{eq:PF_exp_ent}
\hat{\mathbb{E}}[ \mathcal{H}( \hat{b}_{t})] = -\hat{\eta }_{t}\sum _{m=1}^{M}\sum _{i=1}^{N} Z\left( o_{t}^{m} \mid s_{t}^{i}\right) q_{t-1}^{i}\\ 
\cdot log\left(\frac{Z\left( o_{t}^{m} \mid s_{t}^{i}\right)\sum _{j=1}^{N} T\left( s_{t}^{i} \mid s_{t-1}^{j} ,a_{t-1}\right) q_{t-1}^{j}}{\sum _{i'=1}^{N} Z\left( o_{t}^{m} \mid s_{t}^{i'}\right) q_{t-1}^{i'}}\right), \notag 
\end{gather}

\noindent where $\hat{b}\doteq\{ q^{i},s^{i}\}_{i=1}^{N}$ denotes the belief particles with weights $q^{i}$; $M,N$ are the number of observation and state samples accordingly and $\hat{\eta }_{t} =\frac{1}{\sum _{m=1}^{M}\sum _{i=1}^{N} Z\left( o_{t}^{m} \mid s_{t}^{i}\right) q_{t-1}^{i}}$. See appendix A.5 for the full derivation, and \cite{Boers10fusion} for a discussion about convergence of the differential entropy estimator to the true differential entropy value. The estimator for the expected entropy of $\hat{b}_{t}$, \eqref{eq:PF_exp_ent}, is also a function of $\hat{b}_{t-1}$, hence the reward structure $R(\hat{b}, a, \hat{b}')$.

Similar to the discrete case, we use an abstract observation model \eqref{eq:abst}, where instead of discrete observations, summation is done over observation samples, see Figure \ref{fig:abs}. We obtain upper and lower bounds that resemble the results \eqref{eq:lemma_state_theoretical}, \eqref{eq:mean_entropy_bounds} and \eqref{eq:thm_exact} but depend on the approximate expected reward value. Combining results on the expected state-dependent reward and expected entropy,

\begin{theorem} \label{thm2}
The estimated expected reward is bounded by,
\begin{equation} \label{eq:thm_est}
0\leq \!\hat{\bar{\mathbb{E}}}_{o}\left[R\left( \hat{b},a,\hat{\bar{b}}'\right)\right] \!-\!\hat{\mathbb{E}}_{o}\left[R\left( \hat{b},a,\hat{b}'\right)\right] \leq \omega_2 log( K) .
\end{equation}
\end{theorem}
\begin{proof}
see appendix A.6.
\end{proof}
\noindent Here, $\hat{\bar{b}}'\doteq\{ \bar{q}^{i},s^{i}\}_{i=1}^{N}$ denotes a particle set with abstract weight, $\bar{q}^{i}$, due to the abstract observation model \eqref{eq:abst}, and
\begin{equation}
\hat{\bar{\mathbb{E}}}_{o}\left[\cdot\right] \doteq \sum_{m=1}^{M} \sum_{i=1}^{N} \bar{Z}(o^{m} \mid s^{i}) q_{t-1}^{i}\left[\cdot\right] .
\end{equation}
$\hat{\mathbb{E}}_{o}[\cdot]$ and $\hat{b}'$ are defined similarly by replacing the abstract model with the original one. As a result of Theorem \ref{thm2}, the estimated value function is bounded, 
\begin{corollary} \label{corrEstValueFunc}
The difference between the estimated value function and the abstracted value function bounded by, 
\begin{equation} \label{est entropy bounds}
0 \leq \hat{\bar{V}}^{\pi }( b_{t}) -\hat{V}^{\pi }( b_{t}) \leq \mathcal{T}\cdot \omega_2 log( K).
\end{equation}
\end{corollary}
\begin{proof}
see appendix A.7.
\end{proof}
\noindent The computational complexity of the expected reward in \eqref{eq:thm_est} is dominated by complexity of the expected entropy, \eqref{eq:PF_exp_ent}, which is $O(M N^2)$. By choosing $K=M$ the computational complexity of the abstract expected reward diminishes to $O(N^2)$ with bounded loss.

\section{Algorithms} \label{alg_sec}
Since the derivations in previous sections are agnostic to which algorithm is being used, we begin this section by presenting the contribution of our work to an existing algorithm. Then, based on insights gained from the examined algorithm, we propose modifications to improve the current algorithm. In the following section, we show that the changed algorithm empirically surpasses the current SOTA in performance throughout our experiments by a significant margin.

\subsection{Baseline Algorithms}
Sparse sampling (SS) algorithm, introduced in \cite{Kearns02ml}, provides $\epsilon$-accuracy guarantee on the solution at a finite time. However, since it searches the tree exhaustively, the convergence is quite slow in practice. On the other hand, MCTS algorithm \cite{Browne12ieee} has the desirable property of focusing its search on the more promising parts of the tree, but was shown to have poor finite-time performance, requiring an $exp(exp(...exp(1)...))$\footnote{A composition of D-1 exponentials, where D denotes tree depth.} iterations in the worst-case scenario \cite{Munos2014book}. To combat the worst-case running time of MCTS and the slow running time of SS, Forward Search Sparse Sampling (FSSS) \cite{Walsh10aaai} was introduced. It was shown to achieve comparable performance to MCTS with performance guarantees under finite computational budget as in SS.

\subsection{FSSS with Information-Theoretic Rewards}
In its original version, FSSS introduced lower and upper bounds on the estimate of the $Q^d(s,a)$ function. In contrast to SS, FSSS builds the tree incrementally, where each iteration begins at the root node and proceeds down to horizon $H$, to obtain an estimate for the action-value function. Whenever a new node is expanded, its direct action nodes are created alongside $M$ randomly sampled children for each of them. The branching factor, $M$, is a predefined hyper-parameter. After performing $(|A|\cdot M)^d$ iterations, FSSS builds the same tree as SS but enjoys anytime properties. Moreover, FSSS may benefit from reduced computation by utilizing upper and lower bounds and pruning sub-optimal actions. \\
When the reward is defined as an information-theoretic function, such as differential entropy, upper and lower bounds on the value function cannot be determined a-priori, thus no pruning can be made. Nonetheless, our experimental evaluations suggest that FSSS still serves as a strong baseline.

\subsection{Adaptive Information-FSSS}
We coin our new algorithm Adaptive Information FSSS (AI-FSSS). Given the same number of iterations, the actions obtained by the two algorithms are identical. The pseudo-code is presented in appendix B. As in FSSS, we build the tree incrementally, where each iteration adds a new trajectory to the tree. The algorithm constructs an abstract belief tree, where a set of $K$ posterior beliefs share the same reward upper and lower bounds relating it to the underlying reward value. This is done by immediately sampling $K$ observation samples whenever a new action node expanded, followed by a computation of the abstract reward. Based on Theorem \ref{thm2} we derive an \textsc{AdaptBounds} procedure, that adapts the number of aggregated observations. Bounds adaptation halts whenever the highest lower bound, $\underset{a}{\max}  LB(b_{init}a)$, is higher than the upper bound of any other action. This results in the same action selection for the full FSSS and our adaptation, AI-FSSS.

\subsection{Introducing Rollouts to AI-FSSS}
A direct adaptation of FSSS to AI-FSSS would abstract $K$ observations in each new action node up to the full depth of the tree, $d_{max}$. However, when the time budget is limited, it might not be the best strategy, since the abstraction of deeper belief nodes of the tree may never be visited twice, but might need to be refined afterward. Instead, we propose to perform rollout whenever a new action node is met for the first time. This is similar to MCTS, where rollouts are used to get an estimate of the action-value function. This approach will lead to abstraction only for expanded nodes, which are the ones in proximity to the root node. As the number of iterations grows, action nodes are expanded gradually and more abstract belief nodes are added to the tree. Given that the number of iterations equals the number of action nodes in the original Sparse-Sampling tree, followed by \textsc{AdaptBounds} procedure, both algorithms converge to the same solution.

\section{Experiments}

\begin{figure*}[t]
    \centering
    \begin{subfigure}{0.33\textwidth}
         \includegraphics[width=1\textwidth]{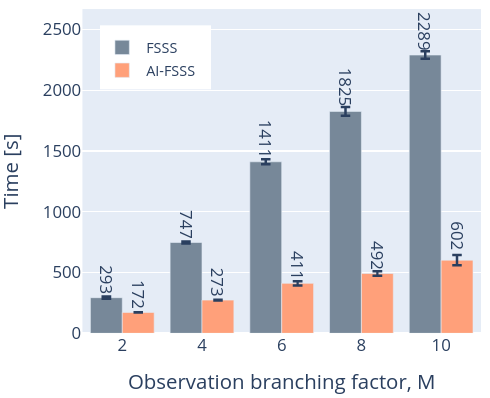}  
         \caption{}
    \end{subfigure}
    \begin{subfigure}{0.33\textwidth}
         \includegraphics[width=1\textwidth]{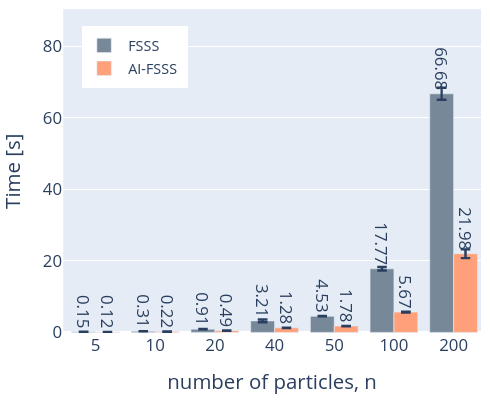} 
         \caption{} 
    \end{subfigure}
    \begin{subfigure}{0.33\textwidth}
         \includegraphics[width=1\textwidth]{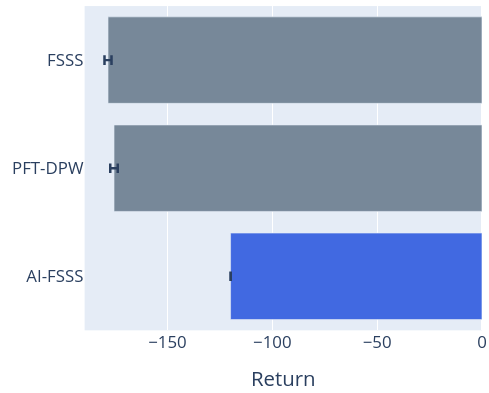} 
         \caption{} 
    \end{subfigure}
    \caption{Evaluating performance of AI-FSSS. (a-b): Running time comparison of FSSS, and our adaptation, AI-FSSS without rollouts. Both algorithms end with the same action selection, but with increasing difference in computation time. (a) Different observation branching factor, with $20$ particles. (b) Different number of state particles, with $4$ observation after each action node. (c) Average total return of AI-FSSS with rollouts in 2D Light-Dark with obstacles.}
    \label{fig:experiments}
\end{figure*}

\begin{figure}[h]
    \centering
    \begin{subfigure}{0.48\textwidth}
         \includegraphics[width=1\textwidth]{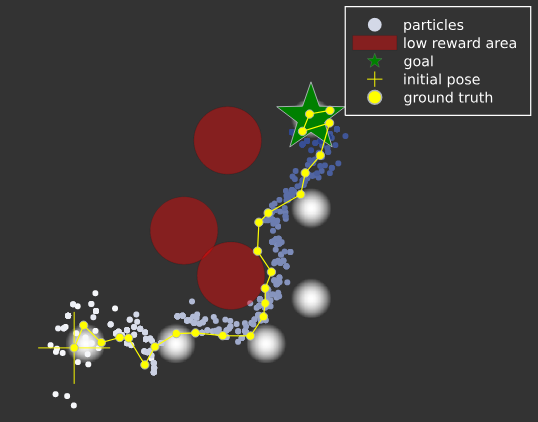}
    \end{subfigure}
    \caption{An illustration of the environment being used in our experiments.}
    \label{fig:Map}
\end{figure}

The goal of the experiment section is to evaluate the influence of the abstraction mechanism on the planning performance. We examined both the time difference and the total return. All algorithms use a particle filter for inference, the choice of the particle filter variant is independent of our contribution. All experiments were performed on the common 2D Light Dark benchmark, where both the state and observation spaces are continuous; see an illustration in Figure \ref{fig:Map}. In this problem, the agent is required to reach the goal while reducing localization uncertainty using beacons scattered across the map. The reward function is defined as a weighted sum of distance to goal, which is state-dependent reward and entropy, as in \eqref{def:reward}. 
Due to space limitations, domain and implementation details are deferred to appendix C.


\subsection{Time Performance Evaluation} 
We compared the basic FSSS with our adaptation, AI-FSSS, in terms of time efficiency. As stressed in previous sections, both algorithms are guaranteed to select the same action. To ensure that both algorithms built the same tree, rollouts were avoided and each iteration proceeded until the maximum depth of the tree. We note that the expected return was inferior to our full algorithm, which is evaluated next. Technically, we also fixed the random numbers by selecting the same seed in both algorithms. 

\noindent \textbf{Observation branching factor.} In the first experiment, we fixed the number of state particles to $n=40$, and examined the influence of different branching factors over the observation space, see Figure \ref{fig:experiments}. The algorithms were limited to 20,000 iterations before performing an action. The computation time indicates an empirical average of over 1,000 simulations, approximating the mean time for a full trajectory. As one would expect, the more observations are clustered in each aggregate, the more time-efficient AI-FSSS compared to the basic FSSS. To obtain the \textit{same} best action in both algorithms, the tree construction was followed by a refinement step, see appendix B, \textsc{AdaptBounds}. In the adapt-bounds step, we incrementally reduce the number of aggregates, so in the worst-case, every aggregate will only hold a single observation and thus recover the FSSS tree. This will occur only in a degenerate case, where all action-values $Q(b_0,\cdot)$ will have the same value, which is rarely the case.

\noindent \textbf{Number of particles.} In our second experiment, we evaluated the effect of the number of particles representing the belief. Here, the number of observations was fixed to $M=4$. Figure \ref{fig:experiments} shows the change in computational speed with regard to the number of particles in our experiments. Both algorithms performed 1,000 simulations; The empirical running time mean and standard deviation are presented in the graph. 
\subsection{Total Return Evaluation}
In the experiments where only few particles were used, e.g. 5, the efficiency gain was mild. In a setting where few particles are sufficient, computing the entropy is relatively cheap compared to other parts of the algorithm, which become relatively more significant (e.g. the different $\max$ operators). However, we observed that the burden became significant even in a mild number of particles, e.g. when $n=20$ the speed-up ratio more than doubled while only a modest cluster size of 4 observations was used.

In contrast to the previous experiments, in this section we evaluate the full version of AI-FSSS, that is, with rollouts for every newly expanded action node. We evaluated performance against FSSS \cite{Walsh10aaai} and PFT-DPW \cite{Sunberg18icaps}, by augmenting a POMDP to a belief-MDP. In this setting, each node holds $n$ particles. 
All algorithms had a 1-second limitation for planning before each interaction with the environment. Except for PFT-DPW, the algorithms shared the same observation branching factor, $M=4$. PFT-DPW opens new observation nodes as time progresses, depending on hyper-parameters defined by the expert. We identified $k_o$ as the dominant hyper-parameter that controls the number of observations in our experiment. We experimented with both $k_o=4$ and $k_o=2$ in order to keep a comparable observation branching size. The better result is shown here, see Figure \ref{fig:experiments}.

Each algorithm performed 1,000 full trajectories in the environment, each containing 25 steps. As suggested in Figure \ref{fig:experiments}, our algorithm performed better than PFT-DPW and FSSS when information-gathering was an explicit part of the task.

\section{Conclusions}
This paper deals with online planning under uncertainty with information-theoretic reward functions. Information-theoretic rewards facilitate explicit reasoning about state uncertainty, contrary to the more common expected reward over the state. Due to the added computational burden of evaluating such measures, we consider an observation model abstraction that improves efficiency. We derived analytical bounds with respect to the original reward function. Additionally, we introduced a new algorithm, AI-FSSS, that contracts the bounds upon need, and is guaranteed to select identical action as the vanilla algorithm. Finally, we conducted an empirical performance study with and without observation abstraction. Our results suggest a significant speed-up as the cardinality of the particle set and the observation-branching factor increases while yielding the same performance.

\section*{Acknowledgements}
The research presented in this paper was partially funded by the Israel Science Foundation (ISF), 
by US NSF/US-Israel BSF, by the Isarel Ministry of Science and Technology (MOST)
and by the Israeli Smart Transportation Research Center (ISTRC).

\bibliographystyle{named}
\bibliography{refs}

\appendix
\newpage
\appendix
\section{Proofs} \label{appendix:appendixA}

\newtheorem{innercustomthm}{Theorem}
\newtheorem{innercustomle}{Lemma}
\newenvironment{customthm}[1]
  {\renewcommand\theinnercustomthm{#1}\innercustomthm}
  {\endinnercustomthm}

\subsection{Lemma 1} \label{proof_lemma1}
The proof is provided for continuous state space; The discrete case obtained similarly by changing integrals to summations.
\begin{proof}
\begin{align}
&\sum _{n=1}^{N_{o}}\bar{\mathbb{P}}\left( o^{n} \mid H^{-}\right)\mathcal{H}( \bar{b}) = \notag\\
-&\sum _{n=1}^{N_{o}}\bar{\mathbb{P}}\left( o^{n} \mid H^{-}\right)\int _{s}\bar{\mathbb{P}}( s\mid H)
\cdot log(\bar{\mathbb{P}}( s\mid H) )
\end{align}
applying Bayes' rule for $\bar{\mathbb{P}}( s\mid H)$,
\begin{align}
-&\sum _{n=1}^{N_{o}}\int _{s}\bar{Z}( o^{n} \mid s) \mathbb{P}\left( s\mid H^{-}\right)\\
\cdot &log\left(\frac{\bar{Z}( o^{n} \mid s) \mathbb{P}\left( s\mid H^{-}\right)}{\int _{s'}\bar{Z}( o^{n} \mid s') \mathbb{P}\left( s'\mid H^{-}\right)}\right) \notag
\end{align}
Splitting summation to follow the partitioning of the abstract observation model,
\begin{align}
-&\sum_{c=1}^{C}\sum _{k=K(c-1)+1}^{Kc}\int_{s}\bar{Z}( o^{k} \mid s) \mathbb{P}\left( s\mid H^{-}\right) \\
\cdot &log\left(\frac{\bar{Z}( o^{k} \mid s) \mathbb{P}\left( s\mid H^{-}\right)}{\int _{s'}\bar{Z}( o^{k} \mid s') \mathbb{P}\left( s'\mid H^{-}\right)}\right) \notag
\end{align}
By construction, $\bar{Z}(o \mid s)$ has uniform distribution for $o^k$, where $k\in[K(c-1)+1,Kc] $. Thus,
\begin{align}
-&\sum_{c=1}^{C} \left[ \sum _{k=K(c-1)+1}^{Kc} 1 \right]\int _{s}\bar{Z}( o^{Kc} \mid s) \mathbb{P}\left( s\mid H^{-}\right)\\
\cdot &log\left(\frac{\bar{Z}( o^{Kc} \mid s) \mathbb{P}\left( s\mid H^{-}\right)}{\int _{s'}\bar{Z}( o^{Kc} \mid s') \mathbb{P}\left( s'\mid H^{-}\right)}\right) = \notag\\
-&\sum_{c=1}^{C} K\cdot \int _{s}\bar{Z}( o^{Kc} \mid s) \mathbb{P}\left( s\mid H^{-}\right)\\
\cdot &log\left(\frac{\bar{Z}( o^{Kc} \mid s) \mathbb{P}\left( s\mid H^{-}\right)}{\int _{s'}\bar{Z}( o^{Kc} \mid s') \mathbb{P}\left( s'\mid H^{-}\right)}\right) = \notag\\
&K\cdot \sum_{c=1}^{C}\bar{\mathbb{P}}\left( o^{Kc} \mid H^{-}\right)\mathcal{H}( \bar{b}) \notag
\end{align}
which concludes the proof.
\end{proof}
\newpage

\subsection{Lemma 2} \label{proof:lemma_state_theoretical}
\begin{proof}
We begin with,
\begin{equation}
    \bar{\mathbb{E}}_o\left[ \mathbb{E}_{s\sim \bar{b}}\left[r\left( s,a\right)\right]\right]
\end{equation}
by definition of $\bar{\mathbb{E}}_o\left[ \cdot \right]$, $\bar{b}(s)$,
\begin{equation}
    \sum _{n=1}^{N_{o}}\bar{\mathbb{P}}\left( o^{n} \mid H^{-}\right)\left[\sum_{s\in S}\bar{\mathbb{P}}(s \mid o^{n}, H^{-})r\left( s,a\right)\right]
\end{equation}
applying chain rule,
\begin{gather}
    \sum _{n=1}^{N_{o}}\sum_{s\in S}\bar{\mathbb{P}}(s,o^{n} \mid H^{-})r\left( s,a\right) = \\ \sum _{n=1}^{N_{o}} \sum_{s\in S} \bar{Z}(o^{n} \mid s) b^-(s) r\left( s,a\right) \notag
\end{gather}
we split the sum over the observations to comply with the abstraction partitioning and use the the abstract observation model definition, (4),
\begin{equation}
    \sum_{s\in S} \sum_{c=1}^{C} \sum_{k=K(c-1)+1}^{Kc} \frac{\sum_{m=K(c-1)+1}^{Kc} Z(o^{m} \mid s)}{K} b^-(s) r\left( s,a\right)
\end{equation}
we then arrive at the desired result,
\begin{gather}
    \sum_{s\in S} \sum_{n=1}^{N_{o}} Z(o^{n} \mid s) b^-(s) r\left( s,a\right) = \\
    \mathbb{E}_o\left[ \mathbb{E}_{s\sim b}\left[r\left( s,a\right)\right]\right]
\end{gather}
\end{proof}
\newpage

\subsection{Theorem 1} \label{proof_thm1}

\begin{proof}
For clarity, we omit the time index in the derivation, the result holds for any time step. We use $H^{-}$ to denote past history while excluding last observation. We also use $b$ and $\mathbb{P}(s\mid o,H^-)$ interchangeably. Rearranging the abstraction from (1), 
\begin{gather*}
     \sum _{k=1}^{K} \bar{Z}\left( o^{k} \mid s\right) = K\cdot \bar{Z}( o^{b}\mid s) \doteq \sum _{k=1}^{K} Z\left( o^{k} \mid s\right)  \ \ \ \ \ \forall b \in [1,K]
\end{gather*}
Plugging it to the expected entropy term, (11),
\begin{align}
&\bar{\mathbb{E}}_{o}\left[\mathcal{H}\left( \bar{b}\right)\right] -\mathbb{E}_{o}\left[\mathcal{H}\left( b\right)\right] = \\
&\sum _{i=1}^{N_o} \bar{\mathbb{P}}\left( o_{i} \mid H^{-}\right)\mathcal{H}\left( \bar{b}\right)
-\sum _{i=1}^{N_o} \mathbb{P}\left( o_{i} \mid H^{-}\right)\mathcal{H}\left( b\right)
\end{align}
expanding the entropy term,
\begin{align} 
-&\sum _{i=1}^{N_o} \bar{\mathbb{P}}\left( o_{i} \mid H^{-}\right)\int _{s} \bar{\mathbb{P}}\left( s \mid o_{i} ,H^{-}\right) log\left( \bar{b}\right) \\ \notag
+ &\sum _{i=1}^{N_o} \mathbb{P}\left( o_{i} \mid H^{-}\right)\int _{s} \mathbb{P}\left( s \mid o_{i} ,H^{-}\right) log\left( b\right)
\end{align}
by Bayes' rule,
\begin{align} 
-&\sum _{i=1}^{N_o}\int _{s} \bar{Z}\left( o_{i} \mid s\right) \mathbb{P}\left( s \mid H^{-}\right) log\left( \bar{b}\right) \\  \notag 
+ &\sum _{i=1}^{N_o}\int _{s} Z\left( o_{i} \mid s\right) \mathbb{P}\left( s \mid H^{-}\right) log\left( b\right)
\end{align}
a change in the order of summation and integral and a split of $N_o = C\cdot K$ result in,
\begin{align} 
-&\int _{s}\sum_{c=1}^{C} \sum _{k=K(c-1)+1}^{Kc} \bar{Z}\left( o_{k} \mid s\right) \mathbb{P}\left( s \mid H^{-}\right) log\left( \bar{b}\right) \\\notag 
+&\int _{s}\sum_{c=1}^{C} \sum _{k=K(c-1)+1}^{Kc} Z\left( o_{k} \mid s\right) \mathbb{P}\left( s \mid H^{-}\right) log\left( b\right)
\end{align}
By plugging-in the definition of the abstract model,
\begin{align} 
-&\int _{s}\sum_{c=1}^{C} \left[\sum _{k=K(c-1)+1}^{Kc} \frac{\sum _{\bar{k}=K(c-1)+1}^{Kc}  Z\left( o_{\bar{k}} \mid s\right)}{K} \right]  \\ \notag
&\mathbb{P}\left( s \mid H^{-}\right) log\left( \bar{b}\right) \\ \notag
+ &\int _{s}\sum _{c=1}^{C}\sum _{k=K(c-1)+1}^{Kc} Z\left( o_{k} \mid s\right) \mathbb{P}\left( s \mid H^{-}\right) log\left( b\right)=\\
-&\int _{s}\sum_{c=1}^{C} \sum _{k=K(c-1)+1}^{Kc} Z\left( o_{k} \mid s\right) \mathbb{P}\left( s \mid H^{-}\right) log\left( \bar{b}\right) \\ \notag
+ &\int _{s}\sum_{c=1}^{C} \sum _{k=K(c-1)+1}^{Kc} Z\left( o_{k} \mid s\right) \mathbb{P}\left( s \mid H^{-}\right) log\left( b\right)=\\
\label{kl_div}
&\sum _{i=1}^{N_o} \mathbb{P}\left( o_{i} \mid H^{-}\right)\int _{s} b \cdot log\left(\frac{b}{\bar{b}}\right)=\\ 
&E_{o}\left[\mathcal{D}_{KL}\left( b ||\bar{b}\right)\right] \geq 0 \notag
\end{align}
\eqref{kl_div} obtained by applying similar steps in reverse order.
The last equality holds since KL-divergence is non-negative and so is its expectation. 
It is left to prove the upper bound; Applying Bayes rule to the nominator and denominator of \eqref{kl_div},

\begin{align}
&\sum _{i=1}^{N_o} \mathbb{P}\left( o_{i} \mid H^{-}\right)\int _{s} b log\left(\frac{Z\left( o_{i} \mid s\right)}{\bar{Z}\left( o_{i} \mid s\right)}\right) \\
+&\sum _{i=1}^{N_o} \mathbb{P}\left( o_{i} \mid H^{-}\right) log\left(\frac{\bar{\mathbb{P}}\left( o_{i} \mid H^{-}\right)}{\mathbb{P}\left( o_{i} \mid H^{-}\right)}\right)\int _{s} b ds \notag
\end{align}
By construction of the abstract observation model, 
\begin{align*}
&\sum_{c=1}^{C} \sum _{k=K(c-1)+1}^{Kc} \mathbb{P}\left( o_{k} \mid H^{-}\right)\int _{s} b \cdot log\left(\frac{Z\left( o_{k} \mid s\right) \cdot K}{\sum\limits _{k=K(c-1)+1}^{Kc} Z\left( o_{\bar{k}} \mid s\right)}\right) ds \\
+&\sum_{c=1}^{C} \sum _{k=K(c-1)+1}^{Kc} \mathbb{P}\left( o_{k} \mid H^{-}\right) log\left(\frac{\bar{\mathbb{P}}\left( o_{k} \mid H^{-}\right)}{\mathbb{P}\left( o_{k} \mid H^{-}\right)}\right)\\
\leq & \ log( K)\sum_{c=1}^{C} \sum _{k=K(c-1)+1}^{Kc} \mathbb{P}\left( o_{k} \mid H^{-}\right)\int _{s} b ds+0=log( K) .
\end{align*}

The inequality is due to positiveness of the denominator in the first term and Jensen's inequality in the second term.
we end up with,

\begin{equation}
0\leq \bar{\mathbb{E}}_{o}\left[\mathcal{H}\left( \bar{b}\right)\right] -\mathbb{E}_{o}\left[\mathcal{H}\left( b\right)\right] \leq log( K) .
\end{equation}
\end{proof}
\newpage

\subsection{Corollary 1.1} \label{proof_corollary}
\begin{proof}
From Lemma 2 it is clear that the expected state-dependent reward is unaffected by the abstraction, and thus will not affect the value function. For the sake of conciseness and clarity, we prove the case that the value function depends only on the entropy. The general case derived similarly by applying Lemma 2 instead of the expected state-dependent reward.
\begin{align}
    V&^{\pi }( b_{t}) = \sum _{n=1}^{N_{o}} \mathbb{P}\left( o_{t+1}^{n} \mid H_{t+1}^{-}\right) \left[ -\mathcal{H}( b_{t+1}) +V^{\pi }( b_{t+1})\right] \notag
\end{align}
expanding the value function,
\begin{align}
    &-\sum _{n=1}^{N_{o}} \mathbb{P}\left( o_{t+1}^{n} \mid H_{t+1}^{-}\right) \bigr[ \mathcal{H}( b_{t+1})\\
    & + \sum _{n'=1}^{N_{o}} \mathbb{P}\left( o_{t+2}^{n'} \mid H_{t+2}^{-}\right)\mathcal{H}( b_{t+2}) +\cdots \bigl]\bigl] \notag
    \end{align}
    by linearity of expectation,
    \begin{align}
    & -\mathbb{E}_{o_{t+1}}\left[ \mathcal{H}( b_{t+1})\right]+ \mathbb{E}_{o_{t+1}}\left[\mathbb{E}_{o_{t+2}}\mathcal{\Bigl[ H}( b_{t+2})\Bigr]\right] +\cdots   \label{use_thm1}
\end{align}
using Theorem 1 for each of the expected entropy terms separately until time-step $\mathcal{T}-1$,
\begin{align}
    V^{\pi }( b_{t}) \geq &-\Bigl[\mathbb{\bar{E}}[\mathcal{H}( \bar{b}_{t+1})] +log( K)\Bigr]\\ 
    &-\mathbb{E}_{o_{t+1}}\left[\bar{\mathbb{E}}_{o_{t+2}}[\mathcal{H}( \bar{b}_{t+2})] +log(K)\right] \cdots  \notag\\
    =& -\bar{\mathbb{E}}_{o_{t+1}}\Bigl[\mathcal{H}( \bar{b}_{t+1})\Bigr]\\
    &-\mathbb{E}_{o_{t+1}}\bar{\mathbb{E}}_{o_{t+2}}\Bigl[\mathcal{H}( \bar{b}_{t+2})\Bigr] \cdots +\mathcal{T}\cdot log(K) \notag
\end{align}
applying similar steps in reverse order yields the abstract value function,
\begin{align}
    &\bar{V}^{\pi }( b_{t}) + \mathcal{T}\cdot log(K) \notag\\
    \notag\\
    \Longrightarrow & \bar{V}^{\pi }( b_{t}) -V^{\pi }( b_{t}) \leq \mathcal{T}\cdot log(K). \notag
\end{align}
Following the same derivation and applying the other side of the inequality of Theorem 1, completes the derivation for the entropy as reward. Using the more general reward, (1), and applying Lemma 2, yields the proof for corollary 1.1,
\begin{equation*}
    0 \leq \bar{V}^{\pi }( b_{t}) -V^{\pi }( b_{t}) \leq \mathcal{T}\cdot \omega_2 log(K).
\end{equation*}
\end{proof}

\newpage
\subsection{Expected Entropy Estimation} \label{exp_ent_est}
We derive an estimator to the expected differential entropy with continuous observation space. The discrete state or observation spaces follows similar derivation by replacing integrals with summations.
\begin{align}
&\mathbb{E}[ \mathcal{H}( b_{t})]
=- \int _{o_{t}} p( o_{t} \mid H_{t}^-)\int _{s_{t}} p( s_{t} \mid o_{t} ,H_{t}^-) \\ & \cdot log( p( s_{t} \mid o_{t} ,H_{t}^-))\notag
\end{align}
applying Bayes' rule,
\begin{align}
\mathbb{E}[ &\mathcal{H}( b_{t})]=-\int _{o_{t}}\int _{s_{t}} p( s_{t} ,o_{t} \mid H_{t}^-) \\ 
& \cdot log\left( Z( o_{t} \mid s_{t})\int _{s_{t-1}} T( s_{t} \mid s_{t-1} ,a_{t-1}) b( s_{t-1})\right)\notag\\
&+\int _{o_{t}}\int _{s_{t}} p( s_{t} ,o_{t} \mid H_{t}^-) \notag \cdot log( p( o_{t} \mid H_{t}^-))\notag
\end{align}
by chain rule and marginalization,
\begin{align}
\mathbb{E}[ &\mathcal{H}( b_{t})]=-\int _{o_{t}}\int _{s_{t}} Z( o_{t} \mid s_{t}) b^{-}( s_{t}) \\ & \cdot log\left( Z( o_{t} \mid s_{t})\int _{s_{t-1}} T( s_{t} \mid s_{t-1} ,a_{t-1}) b( s_{t-1})\right)\notag\\
&+\int _{o_{t}}\int _{s_{t}} Z( o_{t} \mid s_{t}) b^{-}( s_{t}) \notag \\ 
& \cdot log\left(\int _{s_{t}} Z( o_{t} \mid s_{t}) b^{-}( s_{t})\right)\notag
\end{align}
using particle filter, the belief represented as a set of weighted particles, $\{(s^1, q^1),\dots,(s^i, q^i),\dots,(s^n, q^n)\}$. Where $q^i$ denotes the weight of particle $i$.
\begin{align}
\mathbb{E}[ \mathcal{H}( b_{t})]\approx -&\int _{o_{t}} \eta _{t}\sum _{i=1}^{n} Z\left( o_{t} \mid s_{t}^{i}\right) q_{t-1}^{i} \cdot \\ & \cdot log\left( Z\left( o_{t} \mid s_{t}^{i}\right)\sum _{j=1}^{n} p\left( s_{t}^{i} \mid s_{t-1}^{j} ,a_{t-1}\right) q_{t-1}^{j}\right)\notag\\
+&\int _{o_{t}} \eta _{t}\sum _{i=1}^{n} Z\left( o_{t} \mid s_{t}^{i}\right) q_{t-1}^{i} \cdot \notag \\ & \cdot log\left(\sum _{i} Z\left( o_{t} \mid s_{t}^{i}\right) q_{t-1}^{i}\right)\notag
\end{align}
where $\eta _{t} = \int _{o_{t}} \sum _{i=1}^{n} Z\left( o_{t} \mid s_{t}^{i}\right) q_{t-1}^{i}$ normalizes the estimator for the probability function so that it sums to 1. Then, we approximate expectation over the observation space using observation samples, and query the likelihood model conditioned on the state samples, $Z\left( o^{m} \mid s^{i}\right) \ \ \ \forall o^{m} \in \{o^{1},\dots,o^{M}\}$,
\begin{align}
\hat{\mathbb{E}}[ &\mathcal{H}( \hat{b}_{t})] = -\overline{\eta }_{t}\sum _{m=1}^{M}\sum _{i=1}^{n} Z\left( o_{t}^{m} \mid s_{t}^{i}\right) q_{t-1}^{i} \cdot \\ & \cdot log\left( Z\left( o_{t}^{m} \mid s_{t}^{i}\right)\sum _{j=1}^{n} T\left( s_{t}^{i} \mid s_{t-1}^{j} ,a_{t-1}\right) q_{t-1}^{j}\right)\notag\\
+&\overline{\eta }_{t}\sum _{m=1}^{M}\left[\sum _{i=1}^{n} Z\left( o_{t}^{m} \mid s_{t}^{i}\right) q_{t-1}^{i}\right] \notag\\ \notag &  \cdot log\left(\sum _{i'=1}^{n} Z\left( o_{t}^{m} \mid s_{t}^{i'}\right) q_{t-1}^{i'}\right)\notag\\
&\overline{\eta }_{t} =\frac{1}{\sum _{m=1}^{M}\sum _{i=1}^{n} Z\left( o_{t}^{m} \mid s_{t}^{i}\right) q_{t-1}^{i}}
\end{align}
which concludes the derivation.

\newpage
\subsection{Theorem 2} \label{proof_thm2}
Note that the reward function, (1), is built of two terms, state dependent reward and entropy,
\begin{equation}
    R(b,a,b') = \omega_1\mathbb{E}_{s\sim b'}\left[r( s ,a)\right] + \omega_2\mathcal{H}(b').
\end{equation}
For clarity, we divide the proof into two parts, 
\begin{align}
    \!\hat{\bar{\mathbb{E}}}_{o}&\left[R\left( \hat{b},a,\hat{\bar{b}}'\right)\right] \!-\!\hat{\mathbb{E}}_{o}\left[R\left( \hat{b},a,\hat{b}'\right)\right] = \\
    &\omega_1 \left(\hat{\bar{\mathbb{E}}}_{o} \left[\mathbb{E}_{s\sim\hat{\bar{b}}'}\left[r(s,a)\right]\right] -\hat{{\mathbb{E}}}_{o}\left[ \mathbb{E}_{s\sim\hat{{b}}'}\left[r(s,a)\right]\right]\right) \\
    +&\omega_2 \left(\hat{\bar{\mathbb{E}}}_{o}\left[\mathcal{H}\left( \hat{\bar{b}}\right)\right] -\hat{\mathbb{E}}_{o}\left[\mathcal{H}\left( \hat{b}\right)\right]\right) \notag .
\end{align}
The first is about the difference in expected entropy, which is similar in spirit to the proof of Theorem 1. The second follows the claim and proof of Lemma (2). We begin with the difference of the expected entropy. For clarity, we derive the upper and lower bounds separately. For the upper bound,
\begin{proof}
In the following we directly plug-in the expected entropy estimator, with both the abstract observation model and the original observation model. For clarity, we split the expression into two parts and deal with each separately. 
\begin{align}
&\hat{\bar{\mathbb{E}}}_{o}\left[\mathcal{H}\left( \hat{\bar{b}}\right)\right] -\hat{\mathbb{E}}_{o}\left[\mathcal{H}\left( \hat{b}\right)\right] \notag\\
=-&\underbrace{\overline{\eta }_{t}\sum _{m=1}^{M}\sum _{i=1}^{n}\overline{Z}\left( o_{t}^{m} \mid s_{t}^{i}\right) q_{t-1}^{i}}_{(a)}\\
\cdot &\underbrace{log\left(\overline{Z}\left( o_{t}^{m} \mid s_{t}^{i}\right)\sum _{j=1}^{n} T\left( s_{t}^{i} \mid s_{t-1}^{j} ,a_{t-1}\right) q_{t-1}^{j}\right)}_{(a)} \notag\\
+&\underbrace{\overline{\eta }_{t}\sum _{m=1}^{M}\sum _{i=1}^{n}\overline{Z}\left( o_{t}^{m} \mid s_{t}^{i}\right) q_{t-1}^{i} \cdot log\left(\sum _{i=1}^{n}\overline{Z}\left( o_{t}^{m} \mid s_{t}^{i}\right) q_{t-1}^{i}\right)}_{(b)} \notag\\
+&\underbrace{\overline{\eta }_{t}\sum _{m=1}^{M}\sum _{i=1}^{n} Z\left( o_{t}^{m} \mid s_{t}^{i}\right) q_{t-1}^{i}}_{(a)} \notag\\
\cdot &\underbrace{log\left( Z\left( o_{t}^{m} \mid s_{t}^{i}\right)\sum _{j=1}^{n} T\left( s_{t}^{i} \mid s_{t-1}^{j} ,a_{t-1}\right) q_{t-1}^{j}\right)}_{(a)} \notag\\
-&\underbrace{\overline{\eta }_{t}\sum _{m=1}^{M}\sum _{i=1}^{n} Z\left( o_{t}^{m} \mid s_{t}^{i}\right) q_{t-1}^{i} \cdot log\left(\sum _{i=1}^{n} Z\left( o_{t}^{m} \mid s_{t}^{i}\right) q_{t-1}^{i}\right)}_{(b)} \notag
\end{align}
In the first expression we start by splitting the summation to sum over its clusters and sum over the components of each cluster,
\begin{align}
(a)= &\overline{\eta }_{t}\sum _{c=1}^{C}\sum _{k=K(c-1)+1}^{K\cdot c}\sum _{i=1}^{n} Z\left( o_{t}^{k} \mid s_{t}^{i}\right) q_{t-1}^{i} \\ \cdot & log\left(\frac{ Z\left( o_{t}^{k} \mid s_{t}^{i}\right)\sum _{j=1}^{n} p\left( s_{t}^{i} \mid s_{t-1}^{j} ,a_{t-1}\right) q_{t-1}^{j}}{\overline{Z}\left( o_{t}^{k} \mid s_{t}^{i}\right)\sum _{j=1}^{n} p\left( s_{t}^{i} \mid s_{t-1}^{j} ,a_{t-1}\right) q_{t-1}^{j}}\right) \notag\\
(a)= &\overline{\eta }_{t}\sum _{c=1}^{C}\sum _{k=K(c-1)+1}^{K\cdot c}\sum _{i=1}^{n} Z\left( o_{t}^{k} \mid s_{t}^{i}\right) q_{t-1}^{i} \\ \cdot & log\left(\frac{Z\left( o_{t}^{k} \mid s_{t}^{i}\right)}{\overline{Z}\left( o_{t}^{k} \mid s_{t}^{i}\right)}\right) \notag
\end{align}
using the abstract model, (4),
\begin{align}
(a)= &\overline{\eta }_{t}\sum _{c=1}^{C}\sum _{k=K(c-1)+1}^{K\cdot c}\sum _{i=1}^{n} Z\left( o_{t}^{k} \mid s_{t}^{i}\right) q_{t-1}^{i} \\ \cdot & log\left(\frac{K\cdot Z\left( o_{t}^{k} \mid s_{t}^{i}\right)}{\sum _{k=K(c-1)+1}^{K\cdot c} Z\left( o_{t}^{k} \mid s_{t}^{i}\right)}\right).
\end{align}
since the denominator within the log is a sum of positive values, the following clearly holds,
\begin{align}
(a) \leq &\overline{\eta }_{t}\sum _{c=1}^{C}\sum _{k=K(c-1)+1}^{K\cdot c}\sum _{i=1}^{n} Z\left( o_{t}^{k} \mid s_{t}^{i}\right) q_{t-1}^{i} \cdot log( K) 
\end{align} 
by taking the constant $log( K)$ out of the summation, the rest sums to one, so $(a) \leq log( K)$.
Next we bound the second expression from above,
\begin{align}
(b)=&\overline{\eta }_{t}\sum _{c=1}^{C}\sum _{k=K(c-1)+1}^{K\cdot c}\sum _{i=1}^{n} Z\left( o_{t}^{k} \mid s_{t}^{i}\right) q_{t-1}^{i} \\ \cdot & log\left(\frac{\sum _{i=1}^{n}\overline{Z}\left( o_{t}^{k} \mid s_{t}^{i}\right) q_{t-1}^{i} /\overline{\eta }_{t}}{\sum _{i=1}^{n} Z\left( o_{t}^{k} \mid s_{t}^{i}\right) q_{t-1}^{i} /\overline{\eta }_{t}}\right)\notag
\end{align} 
applying Jensen's inequality,
\begin{align}
(b)\leq &log\left(\overline{\eta }_{t}\sum _{c=1}^{C}\sum _{k=K(c-1)+1}^{K\cdot c}\sum _{i=1}^{n}\overline{Z}\left( o_{t}^{k} \mid s_{t}^{i}\right) q_{t-1}^{i}\right)
\end{align}
by recalling the definition of the normalizer, we end up with $log( 1) =0$
\end{proof}
Last, we provide a proof for the lower bound,

\begin{proof}
\begin{align}
&\hat{\bar{\mathbb{E}}}_{o}\left[\mathcal{H}\left( \hat{\bar{b}}\right)\right] -\hat{\mathbb{E}}_{o}\left[\mathcal{H}\left( \hat{b}\right)\right] = \notag \\
-&\overline{\eta }_{t}\sum _{c=1}^{C}\sum _{k=K(c-1)+1}^{K\cdot c}\sum _{i=1}^{n} Z\left( o_{t}^{k} \mid s_{t}^{i}\right) q_{t-1}^{i}\\
\cdot &log\Bigl[\frac{\overline{Z}\left( o_{t}^{k} \mid s_{t}^{i}\right)\sum _{j=1}^{n} T\left( s_{t}^{i} \mid s_{t-1}^{j} ,a_{t-1}\right) q_{t-1}^{j}}{\sum _{i=1}^{n}\overline{Z}\left( o_{t}^{k} \mid s_{t}^{i}\right) q_{t-1}^{i}} \notag\\
\cdot &\frac{\sum _{i=1}^{n} Z\left( o_{t}^{k} \mid s_{t}^{i}\right) q_{t-1}^{i}}{Z\left( o_{t}^{k} \mid s_{t}^{i}\right)\sum _{j=1}^{n} T\left( s_{t}^{i} \mid s_{t-1}^{j} ,a_{t-1}\right) q_{t-1}^{j}}\Bigr] \notag
\end{align}
since $log( x) \leq x-1, \ \forall x >0$,
\begin{align}
&\hat{\bar{\mathbb{E}}}_{o}\left[\mathcal{H}\left( \hat{\bar{b}}\right)\right] -\hat{\mathbb{E}}_{o}\left[\mathcal{H}\left( \hat{b}\right)\right] \geq \notag \\
&\overline{\eta }_{t}\sum _{c=1}^{C}\sum _{k=K(c-1)+1}^{K\cdot c}\sum _{i=1}^{n} Z\left( o_{t}^{k} \mid s_{t}^{i}\right) q_{t-1}^{i}\\
\cdot &\left( 1-\frac{\overline{Z}\left( o_{t}^{k} \mid s_{t}^{i}\right)}{\sum _{i=1}^{n}\overline{Z}\left( o_{t}^{k} \mid s_{t}^{i}\right) q_{t-1}^{i}} \cdot \frac{\sum _{i=1}^{n} Z\left( o_{t}^{k} \mid s_{t}^{i}\right) q_{t-1}^{i}}{Z\left( o_{t}^{k} \mid s_{t}^{i}\right)}\right) \notag
\end{align} 
rearranging terms,
\begin{align}
&1-\overline{\eta }_{t}\sum _{c=1}^{C}\sum _{k=K(c-1)+1}^{K\cdot c}\sum _{i=1}^{n} Z\left( o_{t}^{k} \mid s_{t}^{i}\right) q_{t-1}^{i} \\ \cdot &\frac{\overline{Z}\left( o_{t}^{k} \mid s_{t}^{i}\right)}{\sum _{i=1}^{n}\overline{Z}\left( o_{t}^{k} \mid s_{t}^{i}\right) q_{t-1}^{i}} \cdot \frac{\sum _{i=1}^{n} Z\left( o_{t}^{k} \mid s_{t}^{i}\right) q_{t-1}^{i}}{Z\left( o_{t}^{k} \mid s_{t}^{i}\right)}
\end{align} 
we conclude with,
\begin{align}
&1-\overline{\eta }_{t}\sum _{c=1}^{C}\sum _{k=K(c-1)+1}^{K\cdot c}\sum _{i=1}^{n} Z\left( o_{t}^{k} \mid s_{t}^{i}\right) q_{t-1}^{i} =0
\end{align}
\end{proof}

We now derive the second part of Theorem 2, i.e. for the difference of expected state-dependent reward.

\begin{lemma} \label{lemma_state_PF}
The value of the estimated expected state-dependent reward is not affected by the abstraction shown in (4), i.e.,
\begin{gather}
\hat{\bar{\mathbb{E}}}_{o} \left[\mathbb{E}_{s\sim\hat{\bar{b}}'}\left[r(s,a)\right]\right] =\hat{{\mathbb{E}}}_{o}\left[ \mathbb{E}_{s\sim\hat{{b}}'}\left[r(s,a)\right]\right] \label{eq:lemma_state_PF}
\end{gather}
\end{lemma}

\begin{proof}
\begin{align}
\hat{\bar{\mathbb{E}}}_{o} &\left[\mathbb{E}_{s\sim\hat{\bar{b_{t}}}'}\left[r(s_{t},a_{t})\right]\right] =\\ &\sum _{m=1}^{M} \mathbb{\bar{P}}(o^{m}_{t}\mid H_{t}^-) \sum _{i=1}^{n} \mathbb{\bar{P}}(s^{i}_{t} \mid o_{t}^{m}, H_{t}^-) r(s^{i}_{t},a_{t})
\end{align}
applying chain rule,
\begin{align}
&\sum _{m=1}^{M} \sum _{i=1}^{n} \mathbb{\bar{P}}(s^{i}_{t}, o_{t}^{m}\mid H_{t}^-) r(s^{i}_{t},a_{t}),
\end{align}
then applying chain-rule from the other direction and using the markovian assumption of the observation model,
\begin{align}
&\sum _{m=1}^{M} \sum _{i=1}^{n} \bar{Z}(o_{t}^{m} \mid s^{i}_{t})b_{t}^- \cdot r(s^{i}_{t},a_{t}).
\end{align}
Applying the transition function on particles from $b_{t-1}$, does not alter their weights, therefore we receive the following expression,
\begin{align}
&\sum _{c=1}^{C}\sum\limits _{k=K(c-1)+1}^{K\cdot c}\sum _{i=1}^{n} \bar{Z}\left( o_{t}^{k} \mid s_{t}^{i}\right) q_{t-1}^{i} r(s_{t}^{i},a_{t})
\end{align}
Using (1),
\begin{align}
\sum _{c=1}^{C}\sum _{k=K(c-1)+1}^{K\cdot c}\sum _{i=1}^{n} \frac{\sum\limits _{\bar{k}=K(c-1)+1}^{K\cdot c} Z\left( o_{t}^{\bar{k}} \mid s_{t}^{i}\right)}{K} q_{t-1}^{i} r(s_{t}^{i},a_{t}),
\end{align}
followed by canceling the summation over $k$ with $K$ in the denominator,
\begin{align}
&\sum _{m=1}^{M} \sum _{i=1}^{n} Z(o_{t}^{m} \mid s^{i}_{t})b_{t}^- \cdot r(s^{i}_{t},a_{t}).
\end{align}
We then end up with the desired result,
\begin{equation}
    \hat{\bar{\mathbb{E}}}_{o} \left[\mathbb{E}_{s\sim\hat{\bar{b}}'}\left[r(s,a)\right]\right] =\hat{{\mathbb{E}}}_{o}\left[ \mathbb{E}_{s\sim\hat{{b}}'}\left[r(s,a)\right]\right]
\end{equation}

\end{proof}
To conclude the proofs of Theorem 2, note that,

\begin{align}
    &0\leq \omega_1 \left(\hat{\bar{\mathbb{E}}}_{o} \left[\mathbb{E}_{s\sim\hat{\bar{b}}'}\left[r(s,a)\right]\right] -\hat{{\mathbb{E}}}_{o}\left[ \mathbb{E}_{s\sim\hat{{b}}'}\left[r(s,a)\right]\right]\right) \\
    &+\omega_2 \left(\hat{\bar{\mathbb{E}}}_{o}\left[\mathcal{H}\left( \hat{\bar{b}}\right)\right] -\hat{\mathbb{E}}_{o}\left[\mathcal{H}\left( \hat{b}\right)\right]\right) \leq \omega_2 log(K)
\end{align}
\subsection{Corollary 2.1} \label{proof_corollary_PF}
The proof of 2.1 follows closely to the proof in A.4. Replacing the exact value function with its estimated counterpart from A.6 yields the desired result.

\newpage
\section{AI-FSSS} \label{appendix_AI_FSSS}
In this section we present the main procedures to derive our algorithm. The variables used in Algorithm \ref{alg:AI-FSSS} are $b$, $ba$ and $b'$ which represent a belief node, a predicted belief node, i.e. after performing an action and posterior belief, after incorporating a measurement. $C(\cdot)$ denotes a list of their corresponding children. $a$ and $o_{\{1,...,K\}}$ denotes an action and a list of $K$ sampled observations respectively. $\Bar{P}_{o|s}$ is a list holding the abstract probability values of the measurement model, as in equation (4). $R_{state}(\cdot,\cdot)$ denotes a state-dependent reward function, which may be defined arbitrarily. $\gamma$ denotes the discount factor. $LB, UB$ and $N$ are all initialized to zero. \textsc{Rollout} performs a predefined policy. In our experiments, we chose uniform distribution over all actions for the rollout policy.
Algorithm \ref{alg:Solve} uses $b_{init}$, which represents the initial belief at the root node, $n$ is the number of iterations and $d_{max}$, the maximum depth of the planning tree.

\begin{algorithm}[h]
\caption{AI-FSSS}
\label{alg:AI-FSSS}
\textbf{Procedure}: \textsc{Simulate}(b,d) 
\begin{algorithmic}[1] 
\IF{d = 0}
\RETURN 0, 0
\ELSIF{$|C(b)|<|\mathcal{A}|$}
\STATE $a,o_{\{1,...,K\}} \label{bin} \xleftarrow{}$\textsc{Gen}$(b,K)$
\STATE $\Bar{P}_{o|s} \xleftarrow{}$ \textsc{AbstractObs}$(ba,o_{\{1,...,K\}}) $ \ \ \ \ \ \ \ \ \ \ \ \ // eq.(4)
\STATE $\bar{\mathbb{E}}[\mathcal{R}(ba) ] \xleftarrow{} $\textsc{ExpectedReward}$(b,a,ba,\Bar{P}_{o|s})$
\ELSE 
\STATE $a \xleftarrow{}$ \textsc{SelectAction}$(b)$
\ENDIF
\STATE $lb \xleftarrow{} \bar{\mathbb{E}}[\mathcal{R}(ba) ]$
\STATE $ub \xleftarrow{} \bar{\mathbb{E}}[\mathcal{R}(ba) ] + log(K) $
\IF{$0 < N(ba) < K$}
    \STATE $o \xleftarrow{}$ \textsc{Pop} $(o_{\{1,...,K\}})$
    \STATE $b' \xleftarrow{}$ \textsc{Posterior}$(b,a,o)$
    \STATE $V_{LB}, V_{UB} \xleftarrow{}$\textsc{Simulate}$(b',d-1)$
\ELSIF{$N(ba) = K$} 
    \STATE $b' \xleftarrow{} \underset{b'}{\arg \min}  N(b')$
    \STATE $V_{LB}, V_{UB} \xleftarrow{}$\textsc{Simulate}$(b',d-1)$
\ELSIF{$N(ba) = 0$}
    \STATE $V_{LB}, V_{UB} \xleftarrow{}$\textsc{Rollout}$(ba,d-1)$
\ENDIF
\STATE $LB(ba) \xleftarrow{} lb + \frac{ \gamma V_{LB} + (|C(ba)|-1)( LB(ba) - lb )}{|C(ba)|} $
\STATE $UB(ba) \xleftarrow{} ub + \frac{ \gamma V_{UB} + (|C(ba)|-1)( LB(ba) - ub )}{|C(ba)|}   $
\STATE $LB(b) \xleftarrow{} \underset{a}{\max}  LB(ba)$
\STATE $UB(b) \xleftarrow{} \underset{a}{\max}  UB(ba)$
\STATE $N(b) \xleftarrow{} N(b) + 1$
\STATE $N(ba) \xleftarrow{} N(ba) + 1$
\STATE \textbf{return} $LB(b), UB(b)$
\end{algorithmic}
\end{algorithm}

\begin{algorithm}[h]
\caption{\textsc{Solve}}
\label{alg:Solve}
\textbf{Procedure}: \textsc{Solve} 
\begin{algorithmic}[1] 
\FOR{$i \in 1:n$}
\STATE \textsc{\textsc{Simulate}}$(b_{init},d_{max})$
\ENDFOR
\STATE $action \xleftarrow{}$ \textsc{AdaptBounds}$(b_{init})$
\STATE \textbf{return} $action$
\end{algorithmic}
\end{algorithm}

\begin{algorithm}[h]
\caption{\textsc{Refine}}
\label{alg:Refine}
\textbf{Procedure}: \textsc{Refine}(b,ba,d) 
\begin{algorithmic}[1] 
\IF{IsLeaf(b)}
\RETURN 0, 0

\ELSIF{\textsc{Abstract}$(ba)$}
\STATE $r_{old}  \xleftarrow{} $\textsc{ReuseReward}$(ba)$
\STATE $P_{o|s} \xleftarrow{}$ \textsc{OriginalObsModel}$(ba,o_{\{1,...,K\}}) $
\STATE $\mathbb{E}[\mathcal{H}(ba)]  \xleftarrow{} $\textsc{ExpectedEntropy}$(b,ba,P_{o|s})$
\STATE $r \xleftarrow{} r_{old} + \omega_2 ( \mathbb{E}[\mathcal{H}(ba)]- \bar{\mathbb{E}}[\mathcal{H}(ba)]) $
\ELSE 
\STATE $r  \xleftarrow{} r_{old} $
\ENDIF

\STATE $b' \xleftarrow{} \underset{b'}{\arg \max}  (UB(b') - LB(b'))$
\STATE $a' \xleftarrow{} \underset{a'}{\arg \max}  (UB(b'a') - LB(b'a'))$

\STATE $V_{LB}, V_{UB} \xleftarrow{}$\textsc{Refine}$(b',b'a',d-1)$

\STATE $LB(ba) \xleftarrow{} lb + \frac{ \gamma V_{LB} + (|C(ba)|-1)( LB(ba) - lb )}{|C(ba)|} $
\STATE $UB(ba) \xleftarrow{} ub + \frac{ \gamma V_{UB} + (|C(ba)|-1)( LB(ba) - ub )}{|C(ba)|}   $
\STATE $LB(b) \xleftarrow{} \underset{a}{\max}  LB(ba)$
\STATE $UB(b) \xleftarrow{} \underset{a}{\max}  UB(ba)$
\STATE \textbf{return} $LB(b), UB(b)$
\end{algorithmic}

\label{alg:IsOverlap}
\textbf{Procedure}: \textsc{AdaptBounds}($b_{init}$) 
\begin{algorithmic}[1] 
\WHILE{$\underset{a^+ \in \mathcal{A}}{\max} LB(b_{init}a^+) < \underset{a \in \mathcal{A} \backslash a^+ }{\max} UB(b_{init}a)$}
\STATE $a^* \xleftarrow{} \underset{a \in \mathcal{A}}{\arg \max} LB(b_{init}a)$
\STATE \textsc{Refine}$(b_{init},b_{init}a^*,d)$
\ENDWHILE
\STATE \textbf{return} $a^*$
\end{algorithmic}
\end{algorithm}

\newpage
\section{Implementation Details} \label{appendix_implementation}
\subsection{Domain}
We compared the different algorithms on a two-dimensional Light Dark environment. In this domain, the unobserved state of the agent is its pose, $(X,Y)$, defined relative to a global coordinate frame, located at $(0,0)$. There are 9 possible actions, eight of which has one unit of translation, and they differ from each other by the direction which is equally spaced on a circle, the ninth action has zero translation. We denote the transition model as $x' = f(x,a,w)$. At each time step, the agent receives a noisy estimate of its position as an observation, denoted by $o = h(x,v)$. In our experiments we chose $w$ and $v$ to be distributed according to a Gaussian noise, although in general they may be arbitrary. The reward function defined as the negative weighted sum of distance to goal and entropy,
\begin{equation}
    r(b,a) = -\mathrm{E}_{b}[\| x-x_{g}\|] - \mathcal{H}(b),
\end{equation}
The prior belief assumed to be Gaussian, $b_0 = \mathcal{N}([0, 0],\Sigma_0)$.  
In all our experiments, we employ a receding horizon approach. At each iteration we calculate a solution from scratch and share no information across different time steps.

\subsection{Domain - Total Return Evaluation}
We performed the experiments on a modification of Light Dark 2D and added forbidden regions to the environment. Whenever the agent crosses to a forbidden region, a -10 reward was added to its immediate reward. Also, we added +10 reward whenever the agent reached the goal and stayed there until the episode terminated. 

\subsection{Hyperparameters}
Here we present the hyperparameters used to evaluate the total return performance.
\begin{table}[h]
\centering
\begin{tabular}{rrrr}
\toprule
AI-FSSS \\
\midrule
$n$ & $C$ & $K$ \\
$20$ & $4$ & $4$ \\
\toprule
FFFS \\
\midrule
$n$ & $C$ \\
$20$ & $4$ \\
\toprule
PFT-DPW \\
\midrule 
$n$ & $c$\footnote{} & $k_o$ & $\alpha_o$ \\
$20$ & $1$ & $4$ & $0.014$ \\
\bottomrule
\end{tabular}
\caption{Hyperparameters used in the experiments.\\ \\ $^2$ $c$ controls the bonus of the UCB function, which is different from the observation branching factor in FSSS and AI-FSSS, $C$. }
\label{tab:hyperparams}
\end{table}
\end{document}